\documentclass[11pt,a4paper]{article}
\usepackage[hyperref]{acl2020}
\hypersetup{breaklinks=true}
\usepackage{times}
\usepackage{subfigure}
\usepackage{latexsym}
\usepackage{dsfont}

\newcommand{\gend}{\small{\% gend.}}
\newcommand{\words}{\small{words}}
\newcommand{\pctm}{\small{\% male}}
\newcommand{\bias}{\small{bias}}
\newcommand{\score}{\small{score}}
\newcommand{\ptwo}{{\emph{[P1:]  \space}}}
\newcommand{\lotsospace}{~~~~~~~~~~~}
\newcommand{\allmodel}{\emph{ALL}}
\newcommand{\pone}{{\emph{[P2:]  \space}}}
\usepackage{url}

\aclfinalcopy % Uncomment this line for the final submission

\usepackage{verbatim}
\usepackage{xspace} 
\usepackage{amsmath}
\usepackage{graphicx}
\usepackage{adjustbox}
\usepackage{array}
\usepackage{multirow}
\usepackage{booktabs}
\usepackage{caption}
\usepackage{tipa} % for Tim Rochtashel's diacritic

\newcounter{trCounter}
\newif\iftrvar
\trvartrue
\iftrvar
\newcommand{\tim}[1]{{\small \color{blue} \refstepcounter{trCounter}\textsf{[TR]$_{\arabic{trCounter}}$:{#1}}}}
\else
\newcommand{\tim}[1]{}
\fi

\usepackage{txfonts}

\title{Queens are Powerful too: Mitigating Gender Bias in Dialogue Generation}
\author{
Emily Dinan$^*$, Angela Fan\thanks{~Joint first authors.} \dag, Adina Williams, Jack Urbanek, Douwe Kiela, Jason Weston \\ 
Facebook AI Research \\ 
\dag Laboratoire Lorrain d'Informatique et 
Applications (LORIA)
}
 
\date{}

\begin{document}
\maketitle
\begin{abstract}
Models often easily learn biases present in the training data, and their predictions directly reflect this bias. We analyze gender bias in dialogue data, and examine how this bias is actually amplified in subsequent generative chit-chat dialogue models. We measure gender bias in six existing dialogue datasets, and focus on the most biased one, the multi-player text-based fantasy adventure dataset LIGHT \cite{urbanek2019light}, as a testbed for our bias mitigation techniques. The LIGHT dataset is highly imbalanced with respect to gender, containing predominantly male characters, likely because it is entirely collected by crowdworkers and reflects common biases that exist in fantasy or medieval settings. We consider three techniques to mitigate gender bias: counterfactual data augmentation, targeted data collection, and bias controlled training. 
We show that our proposed techniques mitigate gender bias in LIGHT by balancing the genderedness of generated dialogue utterances and are particularly effective in combination. We quantify performance using various evaluation methods---such as quantity of gendered words, a dialogue safety classifier, and human studies---all of which show that our models generate less gendered, but equally engaging chit-chat responses.
\end{abstract}

\section{Introduction}

Machine learning algorithms learn to model patterns present in training datasets, so data quality affects what they learn. 
Model predictions have been shown to directly reflect harmful societal biases present in training datasets, such as racial bias in sports reports \citep{merullo2019} and political bias in news data \citep{fan2019plainsight}. 
Moreover, biases have been discovered in many NLP tasks, for example, in learned word embeddings \cite{bolukbasi2016man,brunet2018understanding,zhao2019gender}, visual semantic role labeling  \citep{zhao2017men}, natural language inference \citep{he2019unlearn}, abusive language classification \citep{park-etal-2018-reducing}, and coreference resolution \citep{winobias}. 
Although research into bias in NLP is maturing, bias in dialogue utterances has received somewhat less attention \citep{dialoguefairness,sheng-etal-2019-woman,ethicalchallenges}. However, with the rapid development of real-world use-cases for dialogue agents, such as interactive assistants, bias in dialogue models has the very real potential not only to replicate existing social biases, but also to exacerbate them. Dialogue debiasing is thus becoming an increasingly important problem in NLP. In this work, we aim to address this issue by measuring gender bias in dialogue data and mitigating its effects on downstream dialogue generation models.

\begin{table}[t]
\begin{center}
\begin{tabular}{p{9em}p{4em}p{3em}}
\toprule
\multicolumn{3}{l}{\textbf{Gendered word counts in dialogue datasets}} \\
\midrule
Dataset & \% gend. words & \% male bias  \\
\midrule
LIGHT & 0.94 & 73.4 \\ 
Reddit & 1.32 & 69.76 \\ 
Wizard of Wikipedia & 0.076 & 65.9 \\ 
Daily Dialog & 1.02 & 59.04 \\ 
Empathetic Dialogues & 2.07 & 53.45 \\ 
ConvAI2 & 1.28 & 50.05 \\ 
\bottomrule
\end{tabular}
\end{center}
\caption{\textbf{Counts of gendered words} in several dialogue datasets. We report the percent of gendered words (\% gend. words) as well as the percentage of male-gendered words among all gendered words (\% male bias). Datasets are arranged in descending order from most to least male biased. Among these, LIGHT has the most male bias, making it an ideal testbed.
}
\label{table:dialogue_data_comp}
\end{table}

\begin{table*}[t!]
\setlength\tabcolsep{2pt}
\begin{center}
\small
\begin{tabular}{p{8em}p{42em}}
 \toprule
    \multicolumn{2}{l}{\textbf{Persona Example (Original LIGHT Dataset)}} \\ 
    \midrule
    \textit{daughter:} & I spend most of my time doing household chores. I want to find meaning in life. I am energetic and happy. \\
    \midrule
    \textit{chief wife}: & I am the king's chief wife.  Of all the women that he has married, or who are his concubines, I am the principal one. I represent the kingdom of my father, who is the king's biggest ally. My sons are the ones who will most likely become the king after the death of my husband.  \\
    \midrule 
    \textit{women:} & I live with my husband and 4 children in the village. I spend my days washing clothing and cleaning our home. My husband works for the royal army defending out town. \\
    \midrule
    \textit{farmer Bob's wife:} & I am farmer Bob's wife. I like to take care of all our animals. I help Farmer Bob everyday on the farm. \\ %can we move to top o p2? i don't want to move when you're typing lol haha yes!!
    
    \midrule
    \textit{mother:} & I am a mother of eight children. I live with my family in a cottage in the countryside. I spend every day tending to the needs of all of my little ones which can be overwhelming, but I always manage to maintain a pleasing disposition and a happy smile. \\
    \midrule
    \textit{wife:} & I am the wife of a farmer. While I may not be the most attractive woman ever, I am loyal and loving. My husband is a good man, but only seems to stay with me out of duty. \\ 
    \midrule 
    \textit{shady lady:} & I am a shady lady. I work in a tavern, and I am willing to trade sexual favors for money. I have to split the money with the tavernkeeper, so that he will offer me a room to work in. I am beginning to get sick from the ``king's evil", which doctors call syphilis. My future is bleak: madness and death. But this is the only way that I can support myself, so I continue. \\
    \bottomrule
\end{tabular}
\caption{\textbf{Examples of gender biased personas} in  LIGHT. In a review that we conducted in this work, none of these characters were flagged as sexist or offensive. For male examples, see Appendix Table~\ref{table:personaexamples_male}.
\label{table:personaexamples}
}
\end{center}
\end{table*}

Previous work has noted that gender bias is prevalent in many machine learning datasets \cite{stock2017convnets,zhao2017men}, and here we analyzed the gender bias in several existing dialogue datasets (see Table \ref{table:dialogue_data_comp}, and \S \ref{sec:eval_bias} for more discussion). As a testbed for our investigation, we chose the dialogue dataset from the LIGHT text adventure world \citep{urbanek2019light}, because we find it to be significantly more male-biased than other comparable dialogue datasets. Not only is it large enough to train neural chit-chat dialogue models, LIGHT is also interesting, because has multiple potential sources of bias---namely, characters, personas, and dialogues. In the dialogue creation process, crowdworkers were presented with a \textbf{character} (with names such as ``farmer" or ``witch") process, as well as an associated \textbf{persona}---a short textual description for the character.  Supplied with characters and personas, crowdworkers were paired, up and tasked with generating a \textbf{dialogue} between the characters. All dialogues contained within LIGHT are entirely crowdsourced---thus susceptible to reflecting the gender biases of crowdworkers \citep{otterbacher2018, barbosa2019}.  We investigate characters, personas, and dialogues as possible sources of bias in turn in \S\ref{sec:eval_bias}.

\begin{table}[t!]
\setlength\tabcolsep{2pt}
\begin{center}
\small
\begin{tabular}{p{5em}p{18em}}
 \toprule
    \multicolumn{2}{l}{\textbf{Dialogue Example (Original LIGHT Dataset)}} \\ 
    \midrule
    \textbf{wife:}&I was married off by my family about five years ago. I spend my days cooking and cleaning so my husband will have something to eat when he returns from his work and can enjoy a clean home. I love my husband dearly because he works very hard to provide for us. \\
    \midrule
    % \multicolumn{2}{l}{\emph{Self:} wife. I was married off by my family about five years ago. I spend my days cooking and cleaning so my husband will have something to eat when he returns from his work and can ... dress has beautiful frills and small jewels embedded on it.} \\
    % \multicolumn{2}{l}{\emph{Partner:} merchant.} \\ 
    %\midrule
    \textit{merchant:} & What a great day for more money. \\
    %\midrule
    \textit{wife}: &  Oh my. That is some thick dust! \\
    %\midrule 
    \textit{merchant:} & Indeed, it is very old. \\
    %\midrule
    \textit{wife:} & This room is going to take a while to clean. You might want to come back later. \\ 
    %\midrule
    \textit{merchant:} &  It is fine I can set my booth up here. \\
    %\midrule
    \textit{wife:} & With all the foot traffic? \\ 
    %\midrule 
    \textit{merchant:} &  Yes it should be ok. \\
    \textit{wife:} & It doesn't appear that anyone ever comes up here! \\
    \textit{merchant:} & Well they will when they know I am here.\\
    \textit{wife:} & I have my doubts but I'll just go about my cleaning. \\
    \textit{merchant:} & Yea sounds like a good idea. \\
    \textit{wife:} & What is that supposed to mean? \\
    \textit{merchant:} &  I am saying we should both do our jobs. \\
    \textit{wife:} & Don't take that tone with me! \\
    \bottomrule
\end{tabular}%
\caption{{A dialogue from the original LIGHT data. The text for the \textit{wife} persona was crowdsourced.} 
\label{table:dialogueexample}
}
\end{center}
\end{table}%

After measuring gender bias in LIGHT, we then explore three bias mitigation techniques, each of which is either wholly novel, or novel in its application to dialogue: (i) Counterfactual Data Augmentation (CDA) \cite{counterfactualDataSubstitution, zmigrod2019}, (ii) a targeted data collection method, which we refer to as Positive-Bias Data collection, and (iii) Bias Controlled text generation. 
We show that these techniques are most effective in combination, resulting in dialogue models that produce engaging responses with measurably less gender bias and offensive content (see \S \ref{sec:results}). Models and code will be released at \url{parl.ai/projects/genderation_bias}. %\url{[URL- ANONYMIZED]}.

\section{Related Work}\label{sec:relatedwork}

Recently, the NLP community has focused on exploring gender bias in NLP systems \citep{sun2019mitigating}, uncovering many gender disparities and harmful biases in algorithms and text (\citealt{cao2019toward, chang2019bias,  chang2019, costa2019, du2019, emami2019,  garimella2019, gaut2019towards, habash2019,hashempour2019, hoyle2019, kang2019, NLee2019, lepp-2019-pardon, qian2019gender, qian2019reducing, sharifirad2019learning,sharifirad2019using, stanovsky2019evaluating, weapons2016}).  Particular attention has been paid to uncovering, analyzing, and removing gender biases in word embeddings
\cite{basta2019, kaneko2019gender, zhao2019gender, genderneutralwordembeddings, bolukbasi2016man}. This word embedding work has extended to multilingual work on gender-marking \citep{gonen2019does,williams2019, zhou2019}. %Even removing only stereotypical or discriminatory gender information---rather than all gender information---from pretrained word embeddings has been attempted \citep{kaneko2019gender}. % "debiasing method that preserves non-discriminative gender-related information, while removing stereotypical discriminative gender biases from pre-trained word embeddings"
Despite these efforts, many methods for debiasing embeddings remain problematic---i.e., they have only succeeded in hiding word embedding biases as opposed to removing them \cite{gonen2019}---making gender debiasing still an open area of research. 

Despite the relatively ample literature on gender debiasing for word-level representations, very little work has focused on sentence representations \cite{debiasingsentencerepresentations, dialoguefairness, sheng-etal-2019-woman, lee2019exploring}. The majority of sentence debiasing work up until this point foregrounds measuring bias \citep{lee2019exploring, sheng-etal-2019-woman}.
%\citep{sheng-etal-2019-woman, serban2016generative} propose novel ways of measuring bias, some of which incorporate psychological methods for identifying discrimination\citep{lee2019exploring}. 
For example, \citeauthor{dialoguefairness} present a test dataset for dialogue created counterfactually by combining templates and hand-created lists of word pairs; this work shows that models produce less diverse dialogues when prompted with sentences containing words describing individuals from underrepresented groups. Acknowledging the obvious importance of measuring gender bias (see, e.g., \citealt{dialoguefairness}), our dialogue work is novel in that we also propose and compare three methods for directly mitigating it\footnote{To the best of our knowledge, only one other work attempts to gender-debias sentence representations  \citep{li2018towards}; however, it extends a word-embedding post-processing method \citep{bolukbasi2016man} shown to be ineffective at removing gender bias \citep{gonen2019} to sentences. Thus, we take a different tack.}.

\section{Measuring Bias}
 \label{sec:eval_bias}
Before one can mitigate bias, one must first measure it. To determine which dataset to focus on, we initially measured both the amount of gendered words used, and the percent of those which referred to male characters, for six existing dialogue datasets (Table \ref{table:dialogue_data_comp}). Throughout, we compare LIGHT to the other dialogue datasets, and find that it is considerably more biased, which leads us to give LIGHT particular attention in this paper. We primarily address three sources of gender bias in dialogue: (i) imbalance in character genders, (ii) personas (Table~\ref{table:personaexamples}), and (iii) dialogues between characters (Table~\ref{table:dialogueexample}).

 Dialogue research has found that incorporating personas, or personality descriptions that ground a speaker's chat, like \textit{I love fishing}, increases engagingness and improves consistency \cite{zhang2018personalizing,shuster2018engaging,mazare2018training,olabiyi2018persona,li-etal-2016-persona}. However, they can also crystallize gender bias \cite{clark2019don,ethicalchallenges}, propagating it to subsequently generated conversations.  We answer three questions in the context of persona-based dialogue: when creating dialogue datasets, (i) do crowdworkers generate an equal number of male and female characters, (ii) do these characters' personas feature sexism or gender biases, and (iii) are the resulting dialogue utterances biased?

\begin{table}[t]
\setlength\tabcolsep{4pt}
\begin{center}
\small
\begin{tabular}{l cccccc}
 \toprule
    & \multicolumn{4}{c}{\bf \# Characters} & \multicolumn{2}{c}{\bf \# Ref} \\
    & {\it F} & {\it M} & {\it N} &  {\it All}  &{\it F}  & {\it M} \\
    \midrule
    \textbf{\emph{LIGHT}} \\
    Orig Data & 159   & 258  & 1460  & 1877 &  439 &  1238 \\
    Swap Persona  & 336 & 230 & 694 & 1260 & 1419 & 1030  \\
    New Charac. & 151 & 120 & 1448 & 1719 & 357 & 275 \\
    \emph{Total} & 646 & 608 & 3602 & 4856 & 2215 & 2543 \\
    \midrule 
    \textbf{\emph{ConvAI2}} \\ 
    Orig Data & 1109 & 1048 & 4214 & 6371 & 1283 & 1148 \\
    \bottomrule
\end{tabular}
\caption{\textbf{Analysis of gender in LIGHT and ConvAI2}: the original LIGHT dataset contains $1.6\times$ as many male-gendered as female-gendered characters. We compare the original dataset with the dataset obtained after gender-swapping personas and collecting new characters (with new personas). The references column indicates the gender of characters mentioned in the personas. By contrast, ConvAI2 contains a roughly equal number of male and female gendered personas.
\label{table:gender_balance}
}
\end{center}
\end{table}

\paragraph{Bias in Number of Characters.}
We first answer the question: do crowdworkers create an equal number of male and female characters? In addition to LIGHT, we also consider the persona-based dialogue dataset ConvAI2 \cite{zhang2018personalizing}.

To examine gender balance in characters, 
we asked annotators to label the gender of each character in both the LIGHT and ConvAI2 datasets based on the persona (choosing \textit{neutral} if the gender was not explicit). This annotation is possible because many personas include text such as \textit{I am a young woman}, although the majority of personas do not mention an explicit gender. 

We find LIGHT characters to be highly gender imbalanced: in  Table~\ref{table:gender_balance}, we can see that there are over 1.6 times as many male characters as female ones\footnote{We use ``female'' and ``male'' for LIGHT characters -- rather than ``woman" and ``man" -- because some are binarily gendered, but not human.}. It is considerably less gender-balanced than ConvAI2, which has a nearly equal number of male and female gendered personas.\footnote{Note that annotators may widen the gender gap by implicitly assuming genders for ungendered personas.} 

\paragraph{Bias in Personas.}
In addition to the stark underrepresentation of female characters, the medieval setting in LIGHT is likely to encourage crowdworkers to generate dialogues accentuating historical biases and inequalities of the time period \cite{bowman_2010,garcia2017privilege}.  
There is no obligation to recreate historical biases: one can instead use creative license to craft a fun world with gender parity. Therefore, we investigate references to men or women in the text of personas, as another source of bias. To motivate this, take for example, a female persona that contains a gendered reference such as \textit{I want to follow in my \textbf{father}'s footsteps} rather than \textit{in my \textbf{mother}'s}. Using gendered relational nouns  \citep{barker1992, williams2018}, such as \textit{father}, doesn't always signal gender bias, but if female characters are predominantly defined in reference to male characters, it becomes a problem. We count the appearance of gendered words in personas using the list compiled by \citet{genderneutralwordembeddings} and find that men are disproportionately referred to in the personas: there are nearly 3x as many mentions of men than women (see  Table~\ref{table:personaexamples} for examples, and  Table~\ref{table:gender_balance} for counts). 

Qualitatively, LIGHT personas contain many examples that strike us as gender biased (see Table \ref{table:personaexamples}). For example, the character description for \textit{girl} contains the line \textit{I regularly clean and cook dinner}. Gender bias and sexism are clearly present in many dialogue datasets \cite{ethicalchallenges}, but finding a clear way to define sexism (and other kinds of unsafe text), let alone measure it at scale, is very challenging. A simple answer is to rely on annotation where annotators operate under their own, albeit subjective, definition(s) of sexism. To assess the pervasiveness of unsafe content in existing personas, we asked three independent annotators to examine each persona for potentially offensive content. If annotators detected content was `offensive' or `maybe offensive', they were asked to place it in one of four categories---racist, sexist, classist, other---and to provide a reason for their response. Just over 2\% of personas were flagged by at least one annotator, and these personas and the dialogues between these personas were removed from the dataset.

\paragraph{Bias in Human-Generated Dialogue Utterances.}
After uncovering bias in the gender of characters and personas--- qualitatively and in number of gendered words---we go on to examine how those biases may propagate to the dialogues that are created from crowdworkers playing the role of these personas.

First, we count the number of male and female gendered words in the training sets of various dialogue datasets (LIGHT, ConvAI2, Reddit, Wizard of Wikipedia, Daily Dialog, Empathetic Dialogues, and ConvAI2), using the same word list as before \cite{genderneutralwordembeddings}. We use this to calculate the percentage of gendered words out of all words, and the percent male bias, or the percentage of male gendered words among all gendered words. Results are shown in Table~\ref{table:dialogue_data_comp}. LIGHT is the most gender imbalanced dataset among all datasets in this table, with a male bias of 73\%. 

With this in mind, we qualitatively examine the LIGHT dataset and find many biased utterances present in the training data. For example, the \textit{queen} persona adheres to negatively stereotyped gender roles when uttering the line \textit{I spend my days doing embroidery and having a talk with the ladies}. Another character admires a \textit{sultry wench with fire in her eyes}. We see the direct effect of the biased persona on the resultant dialogue (see Table \ref{table:dialogueexample}): for example, a  \textit{wife} persona contains the text \textit{I spend my days cooking and cleaning so my husband will have something to eat when he returns from his work...}, and, in dialogue with a \textit{merchant}, discusses only her cleaning duties. The \textit{merchant} even derisively refers to cleaning as the \textit{wife's} job.

\begin{figure*}[t!]
\centering
\includegraphics[width=1\textwidth]{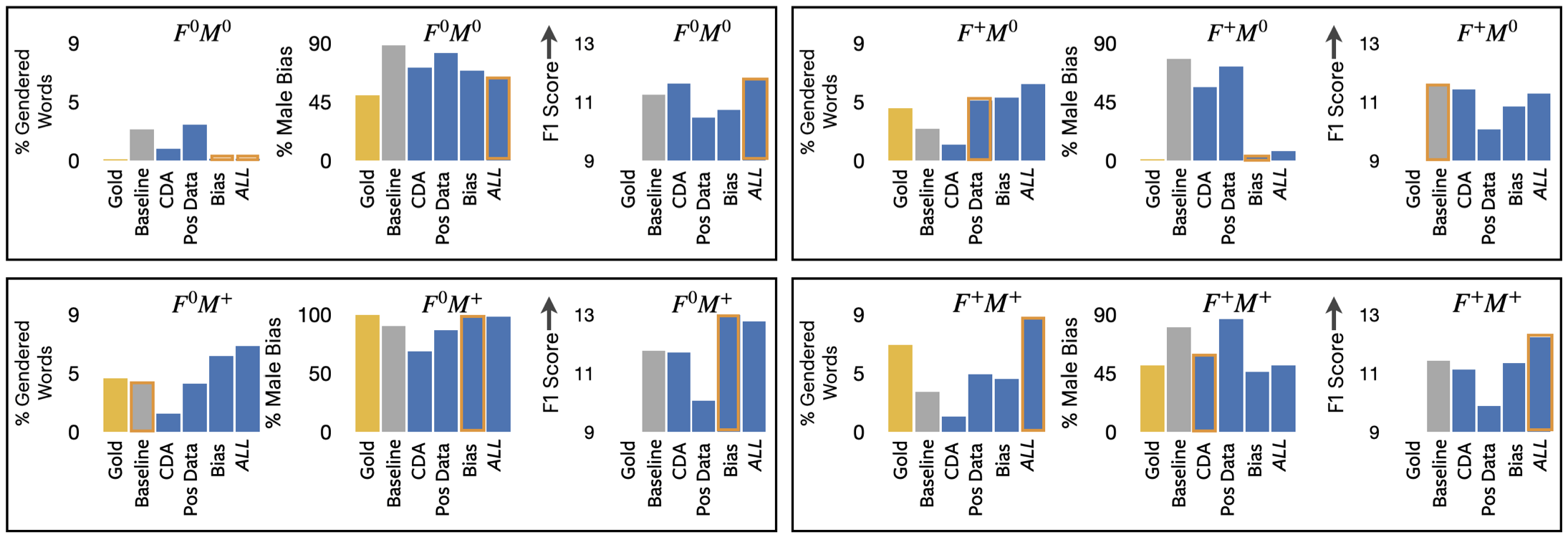}
\caption{We compare the \textbf{performance of various bias mitigation methods}---Counterfactual Data Augmentation (CDA), Positive-Bias Data Collection (Pos. Data), Bias Control Model (Bias Ctrl), and combining these methods (\allmodel)---on the test set, splitting the test set across the four genderedness bins: $\text{F}^{0/+}\text{M}^{0/+}$. $\text{X}^0$ indicates there are no $\text{X}$-gendered words in the gold response, while $\text{X}^{+}$ indicates that there is at least one. We measure the percent of gendered words in the generated utterances (\% gend. words) and the percent of male bias (\% male bias), i.e. the percent of male-gendered words among all gendered words generated. While each of these methods yield some improvement, \textit{combining all of these methods in one yields the best control over the genderedness of the utterances while improving the F1-score.}}
\label{tab:sensitive_f1_results}
\end{figure*}

\section{Mitigating Bias in Generative Dialogue}
As we found the LIGHT was considerably more biased than other dialogue datasets, throughout the rest of the paper we use the LIGHT dataset as a testbed for developing a \emph{general} framework for mitigating bias in generative dialogue. 

When we train dialogue models on biased datasets, the bias will manifest in model-generated dialogues. We explore  data augmentation and other algorithmic methods to mitigate bias in generative Transformer models.  We (i) extend counterfactual data augmentation to dialogue \cite{counterfactualDataSubstitution, zmigrod2019} to swap gendered words, (ii) perform positive data collection by augmenting the existing dataset via targeted data collection with crowdworkers, and lastly,  (iii) present a bias controlled dialogue generation method that controls how many male and female gendered words models produce. 

\subsection{Counterfactual Data Augmentation}\label{sec:cda}

A solution proposed for gender bias in word embeddings is \emph{Counterfactual Data Augmentation} (CDA) \cite{counterfactualDataSubstitution, zmigrod2019, dialoguefairness}. CDA swaps, say, all instances of \emph{grandmother} with \emph{grandfather}, \emph{she} with \emph{he}, and so on. We apply this word-based data augmentation to dialogue generation by first copying every dialogue with a gendered word(s) in it, then swapping it with its pair from the list provided by \citet{genderneutralwordembeddings}. The  augmentation is limited to words on the gendered word list, and the swapping is performed automatically.

\subsection{Positive-Bias Data Collection} 
\label{sec:dataplus}
While CDA has been shown to be a somewhat effective strategy for mitigating bias in word embeddings, this method has several pitfalls: it may result in ungrammatical sentences and it relies on existing (and incomplete) word pair lists to determine and swap gender. To resolve these issues, we use humans to collect additional dialogue data via a two-pronged Positive-Bias Data Collection (Pos. Data) strategy. We first collect additional personas by having humans (i) manually swap the gender of the persona (rather than relying on the word lists) and (ii) write additional, diversified personas. We then use these personas to seed the collection of additional, positively biased dialogue data, which we refer to as Pos. Data throughout.

\paragraph{New Personas.} 

As LIGHT contains more male personas than female personas (see \S \ref{sec:eval_bias}), we balance existing personas with \textbf{gender swapping}. For every gendered persona, annotators create a new opposite-gendered persona for which referring nouns or pronouns are changed, but the rest of the character description remains unchanged. For example, for every persona describing a king, annotators will create a new one describing a queen. Annotators are instructed to swap the gender(s) of other characters referred to in the text (e.g., if an original persona describes a female in relation to her father, the new male persona will describe a male in relation to his mother). This method ensures that the created sentences will be grammatical, unlike heuristic data augmentation.

However, simply balancing references to men and women is insufficient, as female characters might be described in sexist ways (see \S \ref{sec:eval_bias}). As detecting sexism is challenging, we take our qualitative analysis to be sufficient, and move to offset it by collecting a new set of \emph{interesting} and \emph{independent} female characters. We do this by priming workers with examples like \emph{adventurer} with personas like \emph{I am a woman passionate about exploring a world I have not yet seen. I embark on ambitious adventures}. We also provide crowdworkers with additional instruction to guide them towards creating diverse characters: \textit{We're looking for strong and diverse descriptions. Avoid descriptions that could be considered hateful, offensive, or stereotypical}. Even with this explicit instruction, 3 times as many male characters as female characters were created; this fact alone reveals the inherent gender biases of the available crowdworker pool. 
We ultimately exclude all male-gendered personas created in this fashion from the new dataset, which brings  
the number of men and women and the number of references to male or female gendered words to approximate balance in the new dataset (see Table \ref{table:gender_balance}). In total, we add 2,629 new personas.

\begin{figure*}[t!]
\centering
\includegraphics[width=1\textwidth]{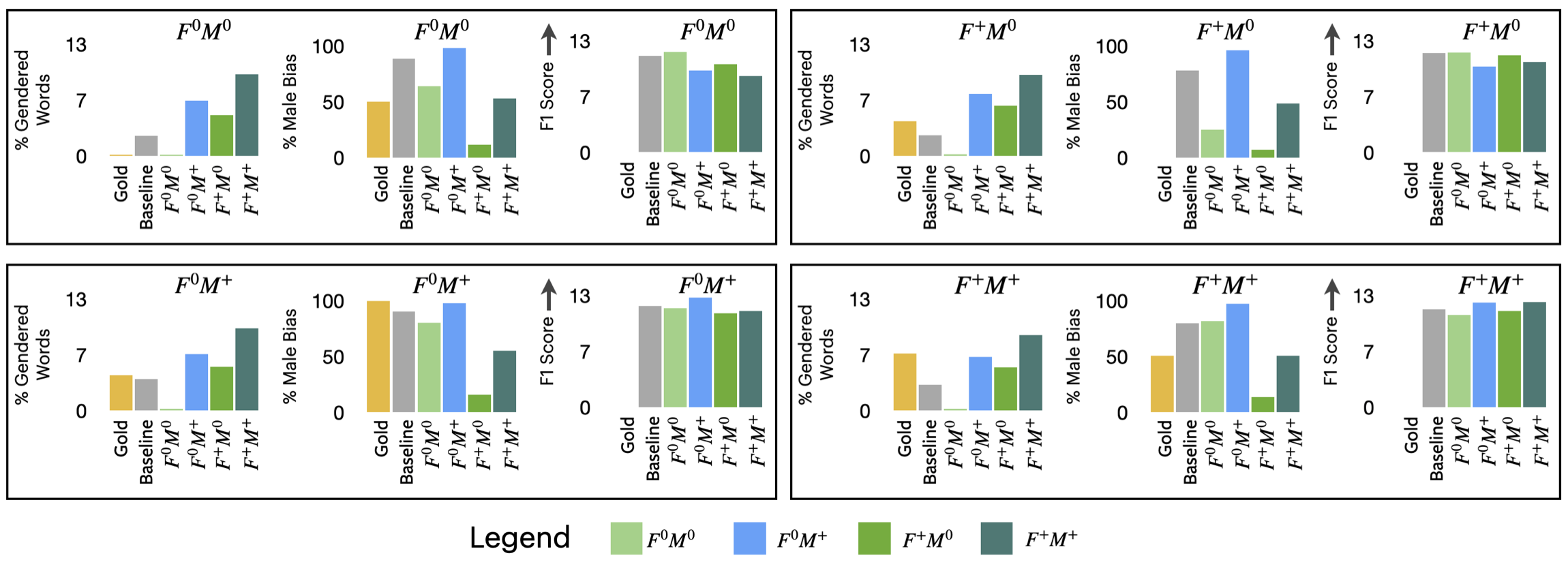}
\caption{\textbf{Performance of the \allmodel~debiasing model} controlled by indicating specific bins for all examples at test time. We report results for each possible conditioning bin choice. Across bins, the model maintains performance as measured by F1 whilst \textit{radically changing the genderedness of the language generated}.}
\label{tbl:forceconditionresults}
\end{figure*}

\paragraph{New Dialogues.}

After gender-balancing the personas, we focus next on our main goal: debiasing generated dialogues. As the personas are a starting point for bias entering the dataset, it is important to address balance in personas as a prior step. 

We use the gender-balanced set of personas derived from the two methods described above to crowdsource additional dialogues. We select more female-gendered characters for new dialogue collection, and instructed annotators to be mindful of gender bias. In particular, we encourage them to assume \emph{equality}---social, economic, political, or otherwise---between genders in this fantasy setting. We collect a total of 507 new dialogues containing 6,658 utterances (approximately ~6\% of the original dataset size). We refer to this additional dialogue data as Pos. Data.

\subsection{Bias Controlled Training}
\label{sec:control}

\begin{table}[t]
\center
\small
\setlength\tabcolsep{6pt} % default value: 6pt
%{\renewcommand{\arraystretch}{1}% for the vertical padding
\begin{tabular}{lrrrrrr}
\toprule
& $\text{F}^{0}\text{M}^{0}$ & $\text{F}^{0}\text{M}^{+}$ & $\text{F}^{+}\text{M}^{0}$ & $\text{F}^{+}\text{M}^{+}$ \\
\midrule 
\% of test set & 60.65 & 27.21 & 7.61 & 4.63 \\ 
\bottomrule
\end{tabular}
\caption{
\textbf{Percentage of dialogue examples in each of the four genderedness bins} ---$\text{F}^{0/+}\text{M}^{0/+}$--- for the LIGHT dialogue data test set.
}
\label{table:binsizes}
%}
\end{table}%

Gender bias in dialogue can take the form of imbalanced use of gendered words. To create dialogue models that can generate an equal number of gendered words, we control model output with Bias Control (Bias Ctrl) via conditional training. Previous conditional training models learn to associate specific control tokens with some desired text properties \cite{kikuchi2016controlling,fan2017controllable,oraby2018controlling,see2019makes}, but have not been applied to address bias issues. 

We apply conditional training techniques to control gender bias in generative dialogue by learning to associate control tokens with properties of gender bias. Any general function that takes as input a dialogue utterance and outputs a continuous or discrete value that provides information about gender bias could be used as a control variable.
In our case, prior to training, each dialogue response is binned into one of four bins--- $\text{F}^{0/+}\text{M}^{0/+}$ ---where $\text{X}^{0}$ indicates that there are zero X-gendered words in the response. $\text{X}^{+}$ indicates the presence of one or more X-gendered word. The percentage of examples from the test set that fall into each bin is noted in Table \ref{table:binsizes}.   Nouns and adjectives are binned via an aggregation of existing gendered word lists  \cite{genderneutralwordembeddings,winobias,hoyle2019}.  Note that other functions could be used as well, such as a bias classifier.  

We append a special token to the input that indicates the bin that the response falls into. During Bias Ctrl training, the model should learn to associate the special token with the genderedness of the dialogue response, such that at inference time, we could modify these special tokens to control the genderedness of the model output. For example, a model trained with multiple gender control bins could be set to the gender neutral (in this case, $\text{F}^{0}\text{M}^{0}$) setting at inference time, to produce a response containing no gendered words.

\subsection{Implementation Details}\label{subsec:models}

Following \citet{urbanek2019light},  we fine-tune a large, pre-trained Transformer encoder-decoder neural network in all generation experiments on the dialogues in the LIGHT dataset. Following \citet{humeau2019real}, the model was pre-trained on Reddit conversations using a previously existing Reddit dataset extracted and obtained by a third party and made available on \url{pushshift.io}. During pre-training, models learned to generate a comment conditioned on the full thread leading up to the comment. Comments containing URLs or  under 5 characters in length were removed, along with child comments, resulting in approximately $2.2$ billion training examples. Similar to pre-training, during fine-tuning, the models are conditioned on the full dialogue history leading up to the next utterance. The model is based on the ParlAI implementation of \citet{miller2017parlai}, and is an 8-layer encoder, 8-layer decoder, with 512 dimensional embeddings and 16 attention heads. For final generations, we decode sequences with beam search size of 5. 

\section{Results}\label{sec:results}

We train five Transformer models: a baseline, three models, one for each of our new methods (see \S\ref{sec:cda} for CDA, \S\ref{sec:dataplus} for Positive-Bias Data Collection, and \S\ref{sec:control} for Bias Control), then one final model, \allmodel, which combines all three methods and achieves the best results. 

\paragraph{Bias is Amplified in Generation.}

Existing Transformer generative dialogue models \cite{serban2016generative,yang2018learning,urbanek2019light} are trained to take the dialogue context as input and generate the next utterance. Generative models are well-known to produce generic text \cite{li2015diversity,fan2018hierarchical}, which makes it likely they will reproduce statistical biases present in datasets. As described previously (see \S \ref{sec:relatedwork}), work shows that machine learning models reflect biases \citep{zhao2019gender,brunet2018understanding}. Moreover, biases can be easier to learn than more challenging reasoning \citep{bolukbasi2016man,lewis2018generative}, suggesting that Transformer models are likely to reflect dataset bias.

Figure~\ref{table:sensitive_f1_results} compares the performance of the various techniques. We compare our methods to the gold labels from the test set and a baseline Transformer generative dialogue model trained on the original data without any bias mitigation techniques. To do this, we divide the test set into four \emph{genderedness bins} (as defined in Section 4.3)---$\text{F}^{0}\text{M}^{0}$, $\text{F}^{0}\text{M}^{+}$, $\text{F}^{+}\text{M}^{0}$, and $\text{F}^{+}\text{M}^{+}$---and calculate: (i) the F1 word overlap with the gold response, (ii) the percentage of gendered words generated (\% gend. words), and (iii) the percentage of male-gendered words generated (relative to the sum total of gendered words generated by the model). 

We find that Transformer models not only reflect dataset  biases, but also they \textit{amplify} them. When the  model produces gendered words (from our gendered word list), it generates male-gendered words the vast majority of the time. Even on utterances for which it is supposed to generate \emph{only} female-gendered words (the gold label only contains female-gendered words), it generates male-gendered words nearly $78\%$ of the time.

\paragraph{Comparing Debiasing Methods} 
As shown in Figure~\ref{tab:sensitive_f1_results}, each of our methods improves the metrics---percent gendered words, percent male bias, and F1---over the baseline Transformer, but we find combining all methods in one in the \allmodel-model is most advantageous. While \allmodel~has more data than CDA and Bias Ctrl, more data alone is not enough --- the Positive-Bias Data Collection model does not achieve as good results. Both the Bias Ctrl and \allmodel~models benefit from knowing the data split ($\text{F}^{0}\text{M}^{0}$, for example), and both yield a gender ratio closest to ground truth. 

\paragraph{Bias Controlled Training Controls Gendered Words.} Our Bias Ctrl method can control the number of gendered words in generated dialogues, as shown in Figure~\ref{table:forceconditionresults}. We examine the effect of Bias Ctrl by generating responses  conditioning the \allmodel~model on each bin. We observe that changing the bin radically changes the genderedness of generated text with only small differences in overall F1. We can control the male bias of the generated dialogue by manipulating these bins.

Examples of generated text from both the baseline and the \allmodel~ model are shown in Table \ref{table:generatedexamples}. The baseline model generates male-gendered words when the gold response contains no gendered words or only female-gendered words, even generating unlikely sequences such as \emph{my name is abigail. i am the king of this kingdom}.

For various methods, we show the top 20 words generated on the test set (after removing stop words) in Table \ref{table:topwords}. We denote gendered nouns using an asterisk. Among the top 20 words generated by the baseline model, there are only two gendered nouns---\textit{knight} and \textit{king}---and both are male-gendered. The 
\allmodel~model generates a similar set of words, but also features \textit{queen} in its top 20 words, another indication that it is more balanced across the male and female genders.

\begin{table}[t]
\setlength\tabcolsep{2pt}
\small
\begin{center}
\begin{tabular}{ll}
\toprule
\multicolumn{2}{l}{ \bf Generation Examples} \\
\midrule
{Bin} & $\text{F}^{0}\text{M}^{0} $ \\ 
{Context} & \ptwo  Owl can you find out how I died? \\
& \pone  I can look around the forest, but I need \\
& \lotsospace more information to help. Tell me what \\ 
& \lotsospace you remember about your past life. \\ 
& \ptwo   I don't remember anything I was hoping \\
& \lotsospace you could find out. \\
& \pone  Your form is very hazy. Do you remember \\ 
& \lotsospace if you are a man or woman? \\ 
{Baseline: } & {\it  i am not a man. i am a man of the forest.}    \\ 
{ALL: } & {\it no, i don't remember. }    \\ 
{Gold: } & {\it  I don't know what's wrong with me!}    \\
\midrule
{Bin} & $\text{F}^{+}\text{M}^{0} $ \\ 
{Context} & \ptwo I do not believe my eyes, for an angel is \\
& \lotsospace upon me!  Angel, please tell me your name.   \\
& \pone  My name is Abigail! \\
{Baseline: } &  {\it my name is abigail. i am the king of this kingdom.} \\ 
{ALL: } & {\it i am the queen's daughter!} \\
{Gold: } & {\it Abigail! Such a beautiful name. To what do I owe} \\
& {\it the pleasure of meeting you?}    \\ 
\bottomrule
\end{tabular}
\end{center}
\caption{\textbf{Example generations} from the baseline model and the proposed debiased models. Gold truth (`Gold') either contains no gendered words or only female-gendered words, but the baseline model still generates male-gendered words.}
\label{table:generatedexamples}
\end{table}

\begin{table}[t!]
\center
\small
\setlength\tabcolsep{6pt} % default value: 6pt
%{\renewcommand{\arraystretch}{1}% for the vertical padding
\begin{tabular}{lrrrrrr}
\toprule
& Gold Labels & Baseline & \allmodel \\
\midrule 
\% Offensive & 13.0 & 14.25 &  {\bf 10.37} \\ 
\bottomrule
\end{tabular}
\caption{
\textbf{Offensive language classification} of model responses on the LIGHT dialogue test set.
}
\label{table:offensive}
%}
\end{table}%

\subsection{Safety of Generated Text} In Table~\ref{table:offensive}, following \newcite{dialoguefairness}, we use a Transformer-based dialogue safety classifier to classify model-generated utterances as offensive or safe. The classifier was fine-tuned on an offensive language classification task \cite{dinan2019build}, and achieves state-of-the-art results. 

We apply this classifier to each utterance generated by the \allmodel~and baseline models on the test set, in addition to the gold (human generated) labels from the test set. Our proposed \allmodel~model is rated as less offensive than both the baseline model and the ground truth (gold) labels (see Table~\ref{table:offensive}).

\subsection{Human Evaluation: Bias and Quality} 
\begin{figure}[t!]
\centering
\includegraphics[width=.27\textwidth]{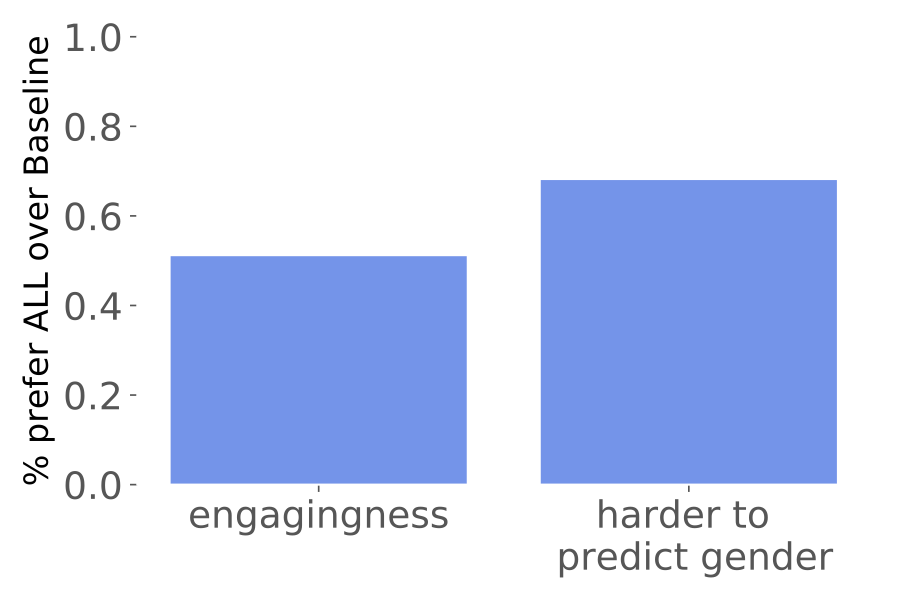}
\caption{\textbf{Human Evaluation of \textit{ALL} model} ($\text{F}^{0}\text{M}^{0}$) compared to baseline Transformer generative model. Evaluators choose which model output they prefer for dialogue engagingness and difficulty of predicting speaker gender. The ALL model produces less gendered text while engagingness is not affected.}
\label{fig:human_eval}
\end{figure}

We compare the quality of our debiasing methods using human evaluation (see Figure~\ref{fig:human_eval}). 
One might hypothesize that some gender debiasing methods work by replacing contentful words (e.g., \textit{witch}) with bleached or uninteresting ones (e.g., \textit{person}, \textit{thing}), effectively trading off gender bias with engagingness. 
We use the dialogue evaluation system Acute-Eval \cite{li2019acute} to ask human evaluators to compare two conversations from different models and decide which model generates (i) more biased dialogues and (ii) more engaging dialogues. We collect 100 model conversations with crowdworkers. Then, we compare conversations between a human and the baseline model to conversations between a human and the \allmodel~model with all generations set to the $\text{F}^{0}\text{M}^{0}$ gender-neutral control bin. Asking for predictions of speaker gender was found to be more effective than asking about sexism or gender bias directly.

As shown in Figure~\ref{fig:human_eval}, it is more challenging to predict the gender of \allmodel~model generations (significant at $p < 0.01$) but the responses are just as engaging according to human evaluators. We conclude our proposed methods are able to mitigate gender bias without degrading dialogue quality. 

\begin{table}[t!]
\small
\center
\setlength\tabcolsep{3pt} % default value: 6pt
{\renewcommand{\arraystretch}{1.2}% for the vertical padding
\begin{tabular}{p{5em}p{17em}}
\toprule
\textbf{Model} & \textbf{Top 20 generated words} \\
\midrule 
Baseline & sorry, hear, not, what, glad, doing, don, king*, thank, sure, will, your, can, much, do, know, but, knight*, blacksmith, going  \\ 
\midrule
\allmodel~ & sorry, hear, sure, not, what, help, doing, your, course, trying, glad, thank, queen*, don, good, king*, but, yes, know, sir*  \\
\midrule
\allmodel~$\text{F}^{0}\text{M}^{0}$ & sorry, hear, sure, what, not, doing, glad, thank, your, yes, course, but, don, do, know, help, have, enjoying, fool, much \\ 
\midrule
\allmodel~$\text{F}^{0}\text{M}^{+}$ & sorry, hear, help, trying, sure, good, king*, sir*, not, your, day, course, father*, he*, don, thank, happy, guard*, glad, have \\
\midrule
\allmodel~$\text{F}^{+}\text{M}^{0}$ & sorry, hear, queen*, sure, miss*, not, your, thank, how, hello, today, guard*, she*, yes, course, kind, woman*, help, glad, what \\ 
\midrule
\allmodel~$\text{F}^{+}\text{M}^{+}$ & sorry, queen*, hear, guard*, help, trying, your, sure, good, course, day, knight*, not, protect, yes, friend, king*, woman*, she*, thank \\

\bottomrule
\end{tabular}
\caption{\textbf{Genderedness bins control the genderedness of generated text}. The top 20 words (test set) with stop words removed. * indicates gendered nouns. 
}
\label{table:topwords}
}
\end{table}

\section{Discussion}\label{sec:error}

\paragraph{Generality of Gendered Words.} The gendered word lists used 
% here 
may not be comprehensive \cite{winobias,genderneutralwordembeddings,hoyle2019}. For example, they do not include \textit{hag} or \textit{wench}, which are 
common in LIGHT. Further, a more continuous representation of gender should be used in the future.

\paragraph{More Fine-Grained Control.} We present an effective method to control the quantity of gendered words generated by manipulating control bins. This technique is general and could be used to control other properties of generated utterances. For example, a sexism or bias classifier could be used instead of the gendered word list.

\paragraph{Quality of Generated Dialogue.} Generative dialogue models are prone to overuse frequent words 
and produce generic utterances, the so-called \textit{I don't know} problem \cite{li2015diversity}. 
We also observe these effects which can affect bias.

\section{Conclusion}

We propose general purpose techniques for reducing gender bias in dialogue. Our methods combine data augmentation, positive-bias data collection, and bias controlled training. Our new data collection techniques help mitigate issues, so clearly bias should be considered at the earliest stages of a project.
Bias control training lessens bias at the training stage, and is also beneficial. Together,
they are especially effective,  
producing less gendered, more gender balanced, safer utterances that maintain engaging dialogue with humans.

\if 0
We propose general purpose techniques for mitigating gender bias in personas and their associated dialogues. Our methods combine data augmentation, positive-bias data collection, and bias controlled training. By manipulating bias control variables, the quantity of gendered words in generated output can be controlled. A combination of our proposed techniques into one \textit{ALL} model reduces bias and produces safer utterances, while still maintaining engaging chit-chat with humans.
\fi

\bibliography{emnlp-ijcnlp-2019}
\bibliographystyle{acl_natbib}

\newpage
\clearpage
%\newcolumn
%\onecolumn
\appendix

\begin{table*}[t!]
\small
\center
\setlength\tabcolsep{3pt} % default value: 6pt
{\renewcommand{\arraystretch}{1.2}% for the vertical padding
\begin{tabular}{lrrrrrrrrrrrrr}
\toprule
\emph{Data Split:} &
\multicolumn{3}{c}{$\text{F}^{0}\text{M}^{0}$} & \multicolumn{3}{c}{$\text{F}^{0}\text{M}^{+}$} & \multicolumn{3}{c}{$\text{F}^{+}\text{M}^{0}$} & \multicolumn{3}{c}{$\text{F}^{+}\text{M}^{+}$} & \multicolumn{1}{c}{All} \\
 \cmidrule(lr){2-4}\cmidrule(lr){5-7} \cmidrule(lr){8-10} \cmidrule(lr){11-13} \cmidrule(lr){14-14}
 & \gend & \pctm & F1 & \gend & \pctm & F1 & \gend & \pctm & F1 & \gend & \pctm & F1 & F1 \vspace{-0.7mm} \\ 
\emph{Model} & \words & \bias & \score &  \words & \bias & \score & \words & \bias & \score & \words & \bias & \score & \score \\
\midrule 
Gold Lbl & 0 & 0 & - & 4.11 & 100 & - & 4.03 & 0 & -  & 6.67 & 50.71 & - & - \\ 
Baseline & 2.37 & 88.39 & 11.24 & 3.66 & 90.26 & 11.77 & 2.44 & 77.99 & \textbf{11.54} & 3.05 & 80.05 & 11.43 & 11.42\\ 
ConvAI2 FT & 0.79 & 71.09 & 7.78 & 1.1 & 78.31 & 7.94 & 1.35 & 51.6 & 8.75 & 1.97 & 67.23 &  8.99 & 7.95\\ 
Reddit Base & 2.18 & 73.68 & 9.93 & 3.03 & 81.78 & 11.54 & 2.81 & 52.99 & 10.99 & 3.94 & 63.16 & 12.61 & 10.57 \\ 
\midrule
CDA & 0.88 & 71.03 & 11.63 & 1.38 & 68.57 & 11.7 & 1.2 & 56.18 & 11.43 & 1.17 & 58.01 & 11.12 & 11.62 \\ 
Pos. Data & 2.76 &  82.44 & 10.46 & {\bf 3.68} & 86.43 & 10.07 & {\bf 4.59} & 72.1 & 10.07 & {\bf 4.43} & 86.5 & 9.88 & 10.44 \\ 
Bias Ctrl & {\bf 0.14} & 68.75 & 10.72 & 5.83 & {\bf 98.08} & 13.01 & 4.8 & {\bf 2.69} & 10.84 & 4.05 & 45.86 & 11.35  & 11.38 \\ 
\allmodel & {\bf 0.14} & {\bf 64.19} & \textbf{11.72} & 6.59 & 97.94 & \textbf{12.77} & 5.84 & 7.13 & 11.28 & 8.81 & {\bf 50.94} & \textbf{12.22}  & \textbf{11.99} \\ 
\bottomrule
\end{tabular}
\caption{
We compare the \textbf{performance of various bias mitigation methods}---Counterfactual Data Augmentation (CDA), Positive-Bias Data Collection (Pos. Data), Bias Control Model (Bias Ctrl), and combining these methods (\allmodel)---on the test set, splitting the test set across the four genderedness bins: $\text{F}^{0/+}\text{M}^{0/+}$. $\text{X}^0$ indicates there are no $\text{X}$-gendered words in the gold response, while $\text{X}^{+}$ indicates that there is at least one. We measure the percent of gendered words in the generated utterances (\% gend. words) and the percent of male bias (\% male bias), i.e. the percent of male-gendered words among all gendered words generated. While each of these methods yield some improvement, \textit{combining all of these methods in one yields the best control over the genderedness of the utterances while improving the F1-score}. 
}
\label{table:sensitive_f1_results}
}
\end{table*}

\begin{table*}[t!]
\small
\center
\setlength\tabcolsep{3pt} % default value: 6pt
{\renewcommand{\arraystretch}{1.2}% for the vertical padding
\begin{tabular}{lrrrrrrrrrrrrr}
\toprule
\emph{Data Split:} &
\multicolumn{3}{c}{$\text{F}^{0}\text{M}^{0}$} & \multicolumn{3}{c}{$\text{F}^{0}\text{M}^{+}$} & \multicolumn{3}{c}{$\text{F}^{+}\text{M}^{0}$} & \multicolumn{3}{c}{$\text{F}^{+}\text{M}^{+}$} & \multicolumn{1}{c}{All} \\
 \cmidrule(lr){2-4}\cmidrule(lr){5-7} \cmidrule(lr){8-10} \cmidrule(lr){11-13} \cmidrule(lr){14-14}
 & \gend & \pctm & F1 & \gend & \pctm & F1 & \gend & \pctm & F1 & \gend & \pctm & F1 & F1 \vspace{-0.7mm} \\ 
\emph{Model} & \words & \bias & \score &  \words & \bias & \score & \words & \bias & \score & \words & \bias & \score & \score \\
\midrule 
Gold Lbl & 0 & 0 & - & 4.11 & 100 & - & 4.03 & 0 & -  & 6.67 & 50.71 & - & - \\ 
Baseline & 2.37 & 88.39 & 11.24 & 3.66 & 90.26 & 11.77 & 2.44 & 77.99 & 11.54 & 3.05 & 80.05 & 11.43 & 11.42\\ 
\midrule
\allmodel~$\text{F}^{0}\text{M}^{0}$  & 0.14 & 64.19 & 11.72 & 0.24 & 80.11 & 11.51 & 0.22 & 25.0 & 11.63 & 0.23 & 81.58 & 10.72 & 11.61 \\ 
\allmodel~$\text{F}^{0}\text{M}^{+}$ & 6.47 & 97.97 & 9.58 & 6.59 & 97.94 & 12.77 & 7.22 & 96.33 & 10.0 & 6.27 & 97.52 & 12.21 & 10.6 \\ 
\allmodel~$\text{F}^{+}\text{M}^{0}$ & 4.77 & 11.66 & 10.27 & 5.12 & 15.84 & 10.94 & 5.84 & 7.13 & 11.28 & 5.03 & 13.64 & 11.23 & 10.57 \\
\allmodel~$\text{F}^{+}\text{M}^{+}$ & 9.53 & 53.34 & 8.89 & 9.6 & 55.35 & 11.19 & 9.42 & 48.65 & 10.5 & 8.81 & 50.94 & 12.22 & 9.79 \\ 
\bottomrule
\end{tabular}
\caption{\textbf{Performance of the \allmodel~debiasing model} controlled by indicating specific bins for all examples at test time. We report results for each possible conditioning bin choice. Across bins, the model maintains performance as measured by F1 whilst \textit{radically changing the genderedness of the language generated}.
}
\label{table:forceconditionresults}
}
\end{table*}

\begin{table*}[t!]
\setlength\tabcolsep{2pt}
\begin{center}
\small
\begin{tabular}{p{8em}p{42em}}
 \toprule
    \multicolumn{2}{l}{\textbf{Persona Example (Original LIGHT Dataset)}} \\ 
    \midrule
    \textit{son:} &  I am spoiled and rich. I enjoy running in the castle. I like hide and seek. \\
    \midrule
    \textit{men:} &  I am an average man in the village. I do what ever work that my King requires me to do. At night, I spend my time in the local pub with my fellow men. \\
    \midrule
    \textit{farmer Bob:} &  I was born in a poor village.  I eat what we grow.  I love being close to the earth. \\ 
    \midrule
    \textit{father:} & I am a role model for my children. I provide for the family with meat and I keep a roof over their heads. I am stability to the family, and keep things together and provide safety to my children. \\
    \midrule
    \textit{husband:} & I try to be good to my wife.  I want to provide for my family.  I try to be strong. \\ 
    \bottomrule
\end{tabular}
\caption{\textbf{Examples of male gender biased personas} written for gendered characters in the LIGHT dataset. 
\label{table:personaexamples_male}
}
\end{center}
\end{table*}

%\section*{Acknowledgements} Liddel, Hila Gonen.

\end{document}